\documentclass[10pt,twocolumn,letterpaper]{article}

\usepackage{iccv}

\usepackage{times}
\usepackage{epsfig}
\usepackage{graphicx}
\usepackage{amsmath}
\usepackage{amssymb}
\usepackage{booktabs}
\usepackage{lipsum} 
\usepackage{multirow}
\usepackage{subcaption}

\usepackage[pagebackref=true,breaklinks=true,letterpaper=true,colorlinks,bookmarks=false]{hyperref}

\iccvfinalcopy % *** Uncomment this line for the final submission

\def\iccvPaperID{123456789}

\ificcvfinal\fi

\begin{document}
\newcommand{\ourmethod}{CCOP}
\newcommand{\ourmethodIntro}{Curricular Contrastive Object-level Pre-training (CCOP)}
\newcommand{\ourmethodFull}{Curricular Contrastive Object-level Pre-training}

\newcommand{\pcurrfull}{Spatial Noise Curriculum Learning}
\newcommand{\pcurrintro}{Spatial Noise Curriculum Learning (SNCL)}
\newcommand{\pcurrshort}{SNCL}

\newcommand{\cselectIntro}{Spatial Curriculum Selection (SCS)}
\newcommand{\cselectFull}{Spatial Curriculum Selection}
\newcommand{\cselectShort}{SCS}

\title{Contrastive Object-level Pre-training with Spatial Noise Curriculum Learning}

\author{Chenhongyi Yang$^{1}$\thanks{Corresponding Author. Email: chenhongyi.yang@ed.ac.uk} \quad\quad  Lichao Huang$^{2}$ \quad\quad Elliot J. Crowley$^{1}$ \\
$^{1}$School of Engineering, University of Edinburgh \\
$^{2}$Horizon Robotics\\}

% Documents
\maketitle
\ificcvfinal\thispagestyle{empty}\fi

\begin{abstract}
The goal of contrastive learning based pre-training is to leverage large quantities of unlabeled data to produce a model that can be readily adapted downstream. Current approaches revolve around solving an image discrimination task: given an anchor image, an augmented counterpart of that image, and some other images, the model must produce representations such that the distance between the anchor and its counterpart is small, and the distances between the anchor and the other images are large. There are two significant problems with this approach: (i) by contrasting representations at the image-level, it is hard to generate detailed object-sensitive features that are beneficial to downstream object-level tasks such as instance segmentation; (ii) the augmentation strategy of producing an augmented counterpart is fixed, making learning less effective at the later stages of pre-training. In this work, we introduce \ourmethodIntro~to tackle these problems: (i) we use selective search to find rough object regions and use them to build an inter-image object-level contrastive loss and an intra-image object-level discrimination loss into our pre-training objective; (ii) we present a curriculum learning mechanism that adaptively augments the generated regions, which allows the model to consistently acquire a useful learning signal, even in the later stages of pre-training. Our experiments show that our approach improves on the MoCo v2 baseline by a large margin on multiple object-level tasks when pre-training on multi-object scene image datasets. Code is available at~\url{https://github.com/ChenhongyiYang/CCOP}.

\end{abstract}
\section{Introduction}
\label{sec:intro}

% 1. visual pretraining is good, ssl is good, better than supervised pretraining
% 2. current ssl methods 
% 3. short comings of the current methods
% 4. first problem
% 5. second problem
% 6. contributions

Imagine that you are given an~\emph{object-level task} such as object detection~\cite{girshick2014rich,ren2015faster,lin2017focal}, or instance segmentation~\cite{he2017mask}, but you do not have enough data or compute resources to warrant training a network from scratch. A popular solution is to take a pre-trained network trained on a existing large-scale dataset, and then use it for the downstream object-level task~\cite{girshick2014rich}. This allows the network to benefit from an existing store of data, helping it converges faster and generalise better than training from scratch. But which pre-trained network should you use?

\begin{figure}[!t]
\centering
    \includegraphics[width=\linewidth]{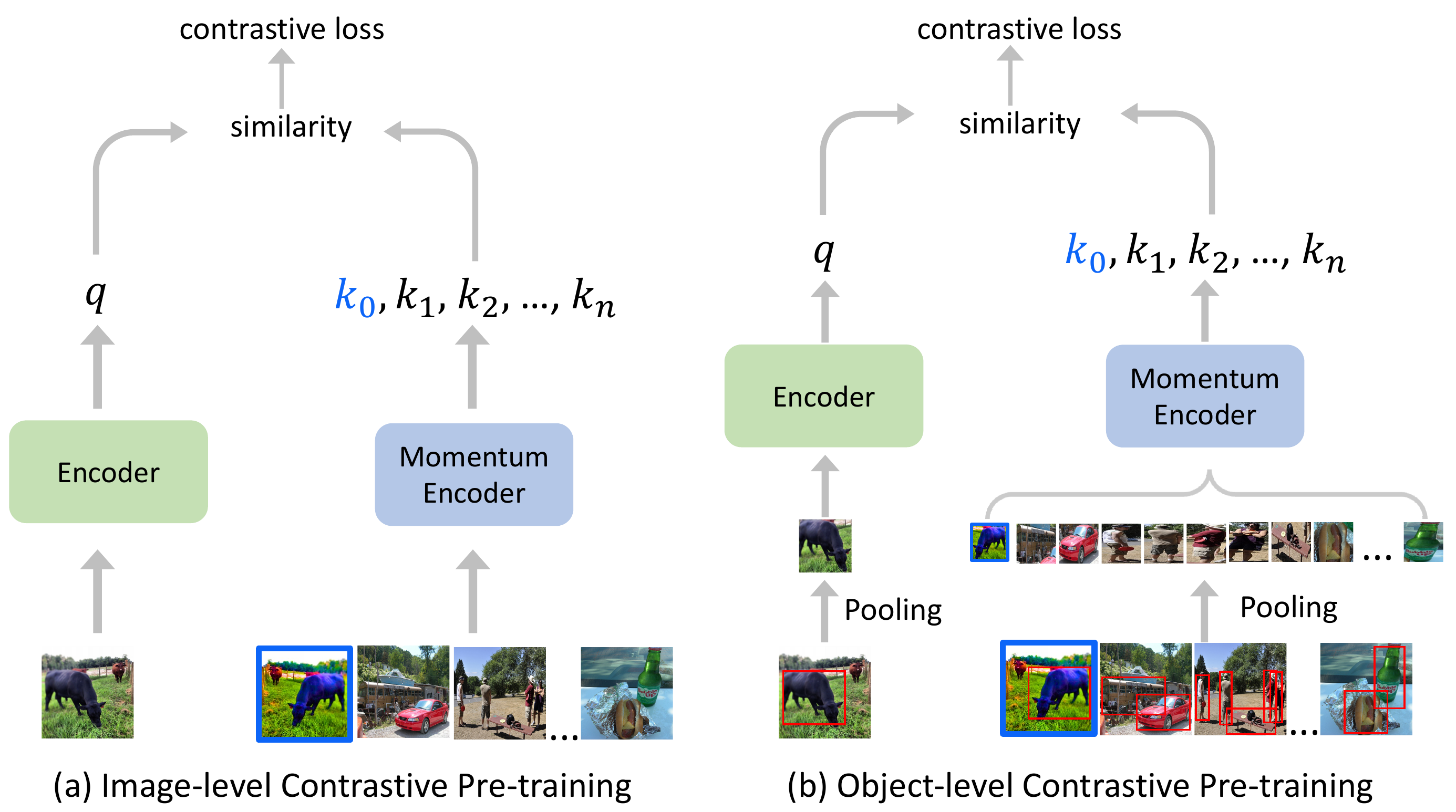}
    \caption{Image-level contrastive pre-training versus object-level contrastive pre-training. The former computes a single representation for each image and uses it in an instance discrimination task. This is not ideal for multi-object scene images. The latter addresses this problem by computing an embedding for each object in the input image and applies object-level instance discrimination.}
    \label{fig:intro}
    % \vspace{-0.5cm}
\end{figure}

For several years, the most popular pre-trained networks were those that had been trained with full supervision on ImageNet~\cite{imagenet}; however, recent works~\cite{simclr,moco,densecl,infomin,detco,detcon,soco,pixpro} have demonstrated that self-supervised~\emph{contrastive learning} can be used to produce comparable, or better networks without requiring any manual annotation. In contrastive learning, a deep model is trained to perform an instance discrimination task, in which two augmented views of an image are produced: the anchor, and its counterpart. Given the anchor, the model must find its counterpart from a huge number of noisy (negative) samples. This is achieved by mapping each image into a low-dimensional embedding space, and using a contrastive loss to maximise the similarity between the anchor image and its positive counterpart while minimising the similarity between the anchor and the noisy samples. Intuitively, this enables the model to discriminate images based on high-level semantic features, rather than low-visual level cues.

However, there are two notable problems with this approach. Firstly, as illustrated in Figure~\ref{fig:intro}, the contrastive loss is applied to distinguish whole images using a global representation, making it hard for the model to produce detailed object-level features; as a consequence, the resulting model may perform poorly when adapted downstream for object-level tasks such as detection and instance segmentation~\cite{ren2015faster,he2017mask}. The second problem lies in how the augmented view of the anchor is built~\cite{infomin}; current approaches use a fixed augmentation strategy throughout the whole pre-training process. However, in the later stages of pre-training, the model is strong enough to easily maximise the similarity between the anchor and its counterpart, saturating the loss gradient provided by the positive pair.

In this work, we present \ourmethodIntro~to address these problems. 
To learn regional spatially-sensitive features we introduce two additional loss terms utilising object-level features; we use selective search~\cite{felzenszwalb2004efficient} to obtain rough object-level bounding boxes, and then extract features through the RoIAlign~\cite{he2017mask} operation. These features are fed into a MLP to produce object-level embeddings. We use a contrastive loss to allow the model to distinguish between object embeddings across different images, and an intra-image discrimination loss to allow the model to differentiate between objects from the same image. These two losses together with a standard image-level contrastive loss enable the model to learn not only global features but also local spatially-sensitive features, and excel when applied downstream to object-level tasks. To solve the loss saturation problem, we introduce a curriculum learning mechanism~\cite{bengio2009curriculum} to make the instance discrimination task harder as pre-training progresses. Specifically, in order to construct positive pairs of a suitable difficulty, we propose \pcurrintro, which uses a \cselectIntro~module that adaptively creates several augmented boxes and selects the most suitable proposal for the loss based on a dynamic difficulty-controlling mechanism. This allows the model to consistently receive a useful learning signal through the whole pre-training process. 

In summary, our contributions are threefold:
\begin{itemize}
\item We present a simple and effective object-level self-supervised pre-training framework called \ourmethodIntro, in which rough object locations are obtained without supervision, and an object-level inter-image contrastive loss and an intra-image discrimination loss are combined to improve pre-training for downstream object-level tasks when training on multi-object scene images.
\item We propose \pcurrintro, which uses a \cselectIntro~module to select the most informative positive pairs at an appropriate difficulty to aid learning. 
\item Through extensive experiments, we show our approach achieves state-of-the-art transfer learning performance on several object-level tasks when pre-training on MS-COCO~\cite{lin2014microsoft}.

\end{itemize}

\section{Related Work}
\label{sec:related}

\begin{figure*}[!ht]
    \centering
    \includegraphics[width=\textwidth]{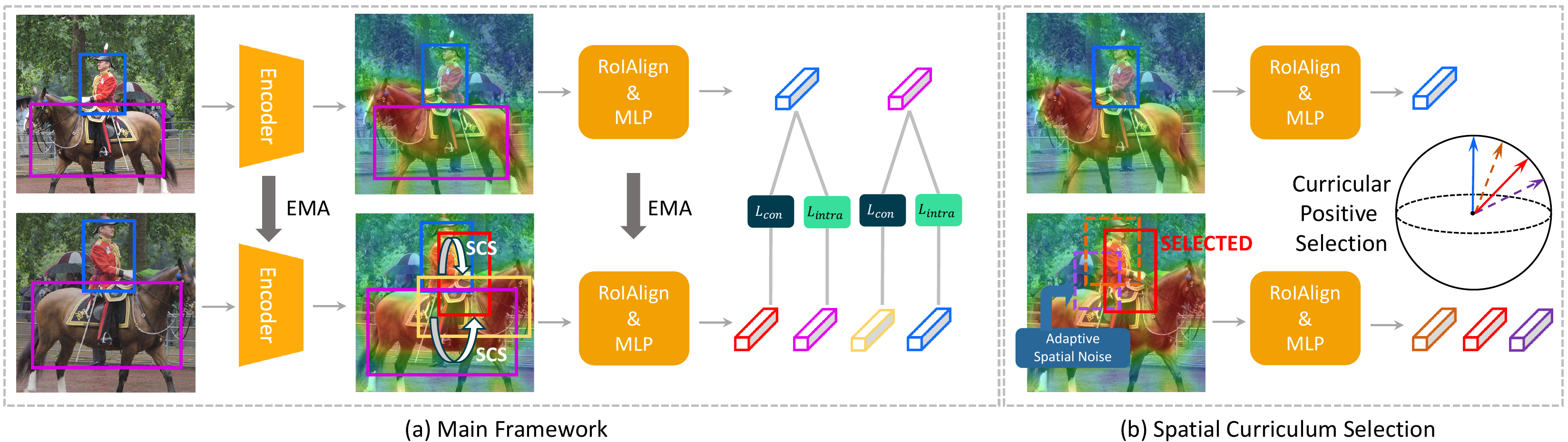}
    \caption{(a) Our main framework. We first compute a set of rough bounding boxes for objects in the input image using selective search. Two different views are fed to the query network (upper branch) and key network (lower branch) respectively. Each box in the key branch is then jittered by our \cselectFull~module to yield augmented boxes of appropriate difficulty, and the hardest of these is selected to be matched with its counterpart in the query branch. (b) In more detail, the Spatial Curriculum Selection module takes a box in the key branch, and produces several jittered boxes where the extent of jittering is proportional to how far we are into pre-training. A feature is computed for each of these boxes, and the feature that is the least similar to the corresponding feature in the query branch is used for learning.}
    \label{fig:pipeline}
    % \vspace{-0.4cm}
\end{figure*}

\subsection{Self-Supervised Pre-training}

Self-supervised methods~\cite{self1, self2, simclr,moco,swav,byol,simsiam} manipulate input data to extract a surrogate supervision signal in the form of a pretext task. There exist numerous pretext tasks including context prediction~\cite{pathak2016context}, grayscale image colourisation~\cite{zhang2016colorful}, jigsaw puzzle solving~\cite{jigsaw}, rotation prediction~\cite{rotation}, and image clustering~ \cite{deep_cluster}.

Recently, instance discrimination as a pretext task has achieved promising results~\cite{simclr}. It considers each image in a dataset as its own class, and uses an InfoNCE loss~\cite{oord2018representation} to match the same instance from among a large number of negative samples. However, this is extremely expensive. MoCo~\cite{moco,mocov2} remedies this by using a memory bank that stores previously-computed representations as negative samples. They rely on noise contrastive estimation to compare instances, which is a special form of contrastive learning. \cite{mocov3} shows that the memory bank can be entirely replaced with elements from the same batch if the batch is large enough. More recently, BYOL~\cite{byol}, SimSiam~\cite{simsiam}, and SWAV~\cite{swav} avoid comparison with negative samples. BYOL and SimSiam directly draw together features from the same instance, while SWAV maps  image features to a set of trainable prototype vectors.

However, models produced through contrasting global image-level representations are unable to capture the inherent differences between image-level representations and object-level, and pixel-level representations. Several works have been proposed to remedy this. In PixPro~\cite{pixpro}, pixel-level contrastive learning is performed by exploiting the geometry between pixels in different views. DetCon~\cite{detcon} uses unsupervised segmentation algorithms to find pixels belonging to the same object and then contrastively learns pixel representations. DenseCL~\cite{densecl} adopts a grid matching mechanism between different images to compute similarity. Self-EMD~\cite{selfemd} improves on this by computing similarity with the earth mover's distance. For object-level pre-training, Ins-Loc~\cite{insloc} combines images to produce pseudo-objects whose regional features are used for contrastive learning. In DetCo~\cite{detco}, a global image representation is composed of several local grid representations so that local features can explicitly contribute to the loss. In SCRL~\cite{scrl}, random regions are selected from images for use in contrastive learning. The closest work to our approach is SoCo~\cite{soco}; this also utilises selective search to find rough objects and forces embeddings of regional representations to be close. However, it uses a fixed strategy for box augmentation, and does not provide a mechanism for the model to distinguish between different objects within an image.

Several works show that constructing good positive and negative samples is essential for contrastive pre-training. In ~\cite{Zhu_2021_ICCV}, hard positive extrapolation is used to increase the distance between positive pairs and negative interpolation is used to improve the diversity of negative samples. A hard-negative mixing mechanism was proposed in~\cite{kalantidis2020hard} to create a hard negative set for each query in MoCo. InfoMin~\cite{infomin} uses an adversarial training strategy to reduce the mutual information between positive pairs.

\subsection{Curriculum Learning}
Curriculum learning was first proposed in~\cite{bengio2009curriculum} to mimic the easy-to-hard learning practices that are commonly used to help with human and animal learning. Recently~\cite{shu2019transferable} used a transferable curriculum learning framework  to allow information transfer between source and target domains; DeepSDF~\cite{duan2020curriculum} was proposed to help reconstruct shapes through gradually increasing difficulty; DCL~\cite{Wang_2019_ICCV_curriculum} uses an adaptive sampling strategy with loss re-weighting to aid learning when given class-imbalanced data. 

Following the core easy-to-hard philosophy of curriculum learning, our proposed \pcurrshort~gradually decrease the similarity between positive object pairs as our model gets stronger.

\section{Method}
\label{sec:method}

In this section we describe our object-level self-supervising training framework: Curricular Contrastive  Object-level  Pre-training (CCOP). It is adapted from MoCo v2, which we review in Section~\ref{sec:moco}. The key innovations of CCOP are (i) the incorporation of object-level objectives, and (ii) spatial noise curriculum learning, which we describe in Sections~\ref{sec:ccop} and Section~\ref{sec:curr} respectively.

\subsection{Description of the Baseline Model}
\label{sec:moco}
Contrastive learning allows models to be trained so that different views of the same image are mapped to be close in some representation space, while different images are mapped to be far away from each other. 

MoCo v2~\cite{mocov2} is a successful contrastive pre-training framework and is adopted as our baseline approach. It comprises a query network $f_q^{\theta_q}$ and a key network $f_k^{\theta_k}$ that share the same architecture, where $\theta_q$ and $\theta_k$ are the model parameters. Each network consists of two parts: a backbone network e.g. ResNet-50~\cite{he2016deep}, and a MLP head that maps the backbone features to an embedding vector. For training, an image $x$ is augmented to produce two views $x_q$ and $x_k$ that are fed into $f_q$ and $f_k$ respectively to get their embeddings $z_q$ and $z_k$, and a memory queue $Q$ is built to store the embeddings generated in previous iterations, which are used as negative samples. A contrastive loss is employed so that the model can match $z_q$ and $z_k$ against a huge number of noisy negative samples $z_k^-$:
\begin{small}
\begin{align}\label{eq:contrastive}
    \mathcal{L}_{con} = -\log{\frac{\exp{(z_q \cdot z_k/\tau)}}{\exp{(z_q \cdot z_k/\tau)} + \sum_{z_k^- \in Q}{\exp{(z_q \cdot z_k^-/\tau)}}}}
\end{align}
\end{small}
where $\tau$ is a temperature hyper-parameter. The query network is updated using the gradient of this loss, and the key network is the exponential moving average (EMA) of the query network:
\begin{align}
    \theta_k \leftarrow m \theta_k' + (1-m) \theta_q.
\end{align}
When pre-training is complete, the backbone of the query network is extracted for downstream use.

\subsection{Object-level Objectives}
\label{sec:ccop}

Our goal is to pre-train a model that will excel when applied to downstream object-level tasks such as instance segmentation and object detection. This pre-training should be conducted without any supervision, so that we are able to leverage large amounts of unlabelled data wherever available in the wild. We would like such a model to produce object-sensitive features. Because of this we choose to use multi-object scene images for pre-training. However, identifying the object regions to produce these features is challenging; this typically requires supervision.

To circumvent this, we take inspiration from~\cite{girshick2015fast} and use selective search~\cite{felzenszwalb2004efficient} to obtain rough bounding boxes for the objects without supervision. We use RoIAlign~\cite{he2017mask} to extract the regional features in a box from the feature map generated by the backbone network. These are fed into a 2-layer MLP to yield object embeddings. Please see Figure~\ref{fig:pipeline} for an illustration. We use a separate memory queue $Q_{obj}$ to store object embeddings. Bounding boxes that correspond to the same object in the two augmented image views are treated as positive pairs, and the regional representations stored in $Q_{obj}$ are used as negative samples in a constrative loss $\mathcal{L}_{con}^{obj}$, which assumes the same form as Equation~\ref{eq:contrastive}.

However, contrasting box embeddings against the negative samples stored in $Q_{obj}$ may allow the model to find a shortcut where it could leverage the global image-level information to avoid learning detailed object-level features. To prevent this from happening, we introduce an intra-image discrimination loss to force the model to compute diverse embeddings for objects in the same image. Given bounding boxes in an image $\{B_{i}\}_{i=1}^{N}$ with corresponding embeddings $\{z_{i}\}_{i=1}^{N}$, we use a hinge loss
\begin{align}\label{eq:imdiscrim}
\mathcal{L}_{intra} = \frac{1}{\left\| \mathbf{P} \right\| _1}\sum_{i=1}^{N}{\sum_{j\neq i}{\mathbf{P}_{ij}\max\{(z_i \cdot z_j-\alpha), 0\}}}.
\end{align}
Here $\mathbf{P}$ is an indicator matrix where $\mathbf{P}_{ij}$ is 1 only if the IoU between $B_i$ and $B_j$ is less than 0.05. Note that we compute $\mathcal{L}_{intra}$ using the original boxes instead of the augmented boxes. This loss reduces the similarity between box embeddings that do not overlap in an image, and can be seen as a form of regularisation which prevents the model from mapping distinct objects close together.  

The final loss function used to train the model is combination of the original image contrastive loss in MoCo v2, the object contrastive loss, and the intra-image discrimination loss:
\begin{align}
\mathcal{L}_{total} = \mathcal{L}_{con}^{img} +  \mathcal{L}_{con}^{obj} +\mathcal{L}_{intra}
\end{align}

\subsection{\pcurrfull}
\label{sec:curr}
In contrastive learning, a fixed data augmentation strategy is often deployed for the whole training process. As a result, the learning signal that the model receives from the contrastive loss will decrease at the later stages of pre-training; as the model becomes stronger it can more easily match positive pairs with a high similarity from the noisy negative samples, which reduces the magnitude of the gradients provided by the positive pair. Let's consider the gradient of $\mathcal{L}_{con}$ with respect to $z_q$. For $\tau = 1$ it is given by
\begin{align}
\nabla_{z_q} \mathcal{L}_{con}= \frac{\sum_{i=1}^{M}{(z^-_i-z_k)\cdot \exp{(z_q \cdot z^-_i)}}}{\exp{(z_q\cdot z_k)} + \sum_{i=1}^{M} \exp{(z_q \cdot z^-_i)}}
\end{align}
where $\{z^-_{i}\}_{i=1}^{M}$ are noisy samples stored in the memory bank, and $M$ is the size of the memory bank (65,335 in our case). In the later stages of training, the $z_q \cdot z^-_i$ terms tend to zero~\cite{Zhu_2021_ICCV}, and this gradient can be approximated as 
\begin{align}\label{eq:approxGradient}
\nabla_{z_q} \mathcal{L}_{con} \approx (z_q - z_k) + \frac{1}{M} \sum_{i}^{M}{(z_{i}^- - z_q)}.
\end{align}
We can see that when $z_k$ is very close to $z_q$, the first term in Equation~\ref{eq:approxGradient} disappears and only negative samples are contributing to the gradient this loss.

To address this problem, we employ curriculum learning to gradually raise the difficulty of the instance-discrimination task so that the model can consistently receive a useful learning signal. Here, we define the difficulty as the similarity between $z_q$ and $z_k$. For image-level contrastive pre-training, one could generate several augmentations and adaptively select the most suitable augmented image for the loss. However, this would be expensive to compute. Fortunately, at the object-level we can augment a bounding box multiple times and quickly acquire their features through the effective RoIAlign operation~\cite{he2017mask}. This faciliates \pcurrfull ~(SNCL).

\begin{figure}[!t]
    \includegraphics[width=\linewidth]{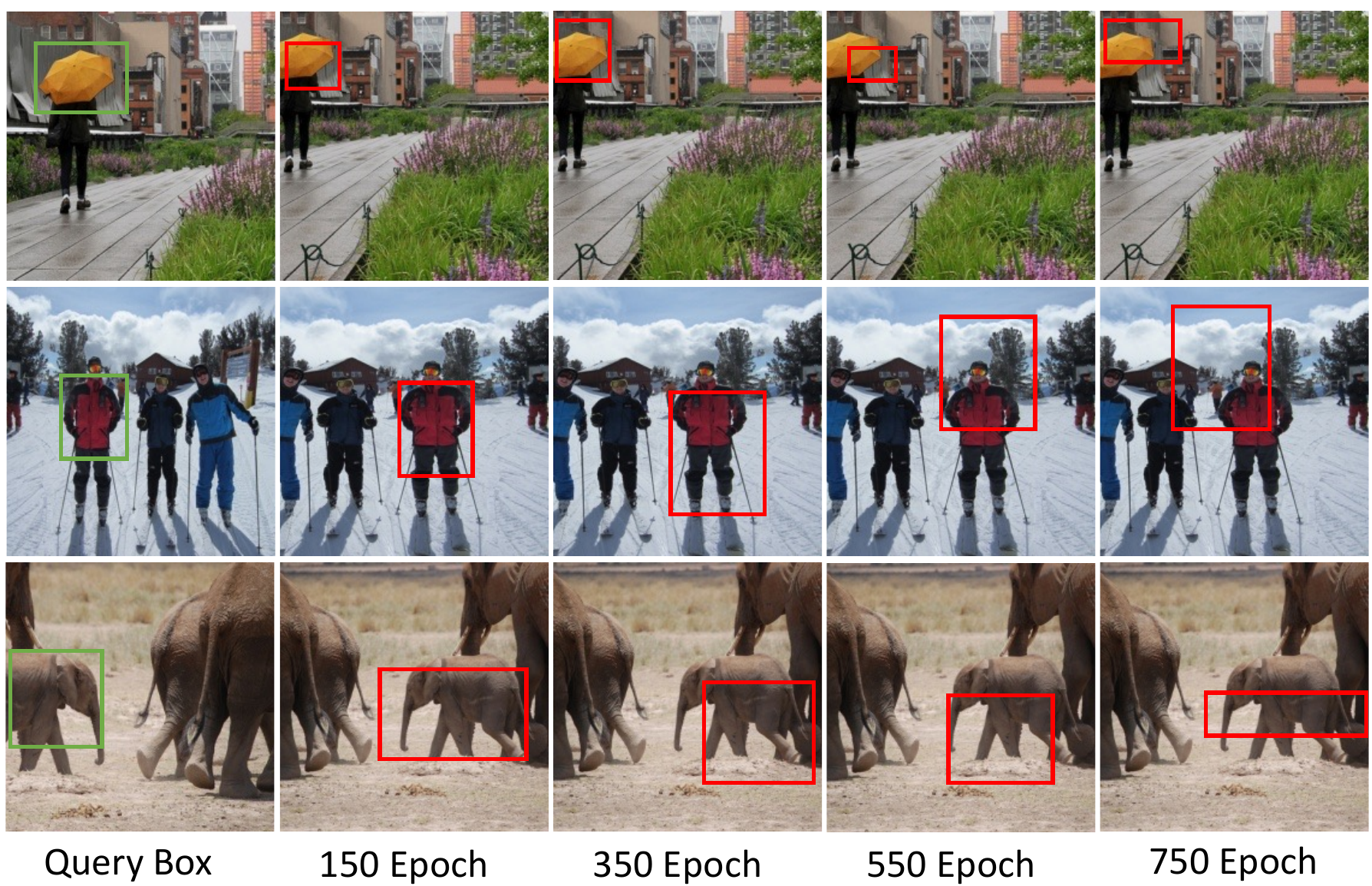}
    \caption{Our \cselectFull ~module augments object bounding boxes so that they are an appropriate difficulty for the model throughout training (note that colour distortions are not shown here).}
    \label{fig:curr}
    % \vspace{-0.5cm}
\end{figure}

\pcurrshort~aims to gradually decrease the similarity of positive object pairs by jittering the bounding boxes as training progresses; we hypothesise that this will allow the model to attend to fine-grained visual cues in order to connect these bounding boxes. In our framework, we achieve this by introducing a \cselectIntro~module, shown in Figure~\ref{fig:pipeline}, that adaptively creates a set of jittered boxes of a dynamically-controlled difficulty, before selecting the hardest of these for learning. In detail, for a box $B^k=(x,y,w,h)$\footnote{$(x,y)$ are the coordinates for the top-left corner of a box with width and height $w$ and $h$ respectively.} in the key branch, the \cselectShort~module generates a set of augmented boxes $\{\hat{B^k_i}\}$ through:
\begin{align}
  &x' = \sigma_x w + x,  \ y'=\sigma_y h + y,\nonumber\\
  &w' = w\cdot \mathrm{exp}(\sigma_w),  \  h' = h\cdot \mathrm{exp}(\sigma_h),
\end{align}
where $\sigma_x,\sigma_y, \sigma_w,\sigma_h$ are noise coefficient drawn from a uniform distribution $(-\zeta, \zeta)$. In order to increase the difficulty, we linearly scale $\zeta$ from 0.3 to 1.0 as training proceeds. We ensure that the IoU between the jittered boxes and the original box is larger than a threshold $\beta_t$, which is dynamically lowered from 0.8 to 0.3 in our \pcurrshort with respect to the training iteration $t$. Then, the \cselectShort~module select the hardest box $\hat{B^k_j}$ to compute $z_k$ for contrastive learning where $j=argmin_{i}\{z_q \cdot \hat{z_i}\}$ . Here $\hat{z_i}$ is the object embedding corresponding to $\hat{B^k_i}$. This enlarges the first term in Equation~\ref{eq:approxGradient}, allowing $z_k$ to contribute to the loss gradient. We visualise the boxes produced by this process for different epochs in Figure~\ref{fig:curr}.
\section{Experiments}
\label{sec:exp}

\subsection{Experimental Setup}
\noindent  \textbf{Datasets.} 
\ourmethod~utilises multi-object scene images as pre-training data to produce a model that excels at object-level downstream tasks. For our experiments we use MS-COCO~\cite{lin2014microsoft} as the pre-training dataset; specifically, the \textit{train2017} subset for pre-training, which contains 118k images. Note that we do not use any annotations.

\noindent \textbf{Model settings.} 
We use MoCo v2~\cite{mocov2} as our baseline. We use ResNet-50~\cite{he2016deep} as the default backbone network. We follow~\cite{soco} to add a feature pyramid network (FPN)~\cite{lin2017feature} to the backbone network, whose generated feature maps are from $\frac{1}{4}$ to $\frac{1}{32}$ of the input image size. Following~\cite{moco} we use the $C_5$ feature map for image embeddings, and the features for object-level embeddings are pooled by a RoIAlign operation from the FPN feature maps according to their size. We use two separate MLPs with intermediate dimensions of 2048 to get 128-D embeddings. Both $Q_{img}$ and $Q_{obj}$ are of size 65,335.

\begin{table}[t]
  \begin{center}
  \resizebox{\linewidth}{!}{
    \begin{tabular}{c|ccc|ccc}
    Method & AP$^{bb}$ & AP$^{bb}_{50}$ & AP$^{bb}_{75}$ & AP$^{mk}$ & AP$^{mk}_{50}$ & AP$^{mk}_{75}$ \\
    \hline
    Rand Init & 31.0 & 49.6 & 33.3 & 28.6 & 46.7 & 30.5 \\
    IN Sup& 38.7 & 59.4 & 42.2 & 35.2 & 56.2 & 37.6 \\
    \hline
    SimCLR~\cite{simclr} & 38.0 & 57.3 & 41.5 & 34.3 & 54.5 & 36.6 \\
    BYOL~\cite{byol} & 38.0 & 57.4 & 41.5 & 34.5 & 54.6 & 36.9  \\
    SimSiam~\cite{simsiam} & 38.3 & 58.0 & 41.8 & 34.8 & 55.2 & 37.2  \\
    MoCo v2~\cite{mocov2} & 38.5 & 58.0 & 41.9 & 34.9 & 55.2 & 37.3 \\
    Self-EMD~\cite{selfemd} & 39.3 & {\bf 60.1} & 42.8 & - & - & - \\
    DenseCL~\cite{densecl} & 39.6 & 59.3 & 43.3 & 35.7 & {\bf56.8} & 38.4 \\
    \hline
    \ourmethod~(Ours) & {\bf 39.9} & 59.8 & {\bf43.9} & {\bf36.2} & {\bf56.8}& {\bf38.8}  \\
    \end{tabular}}
  \end{center}
  \vspace{-0.4cm}
  \caption{COCO Object detection and instance segmentation using Mask R-CNN FPN with 1$\times$ schedule. `Rand Init' refers to training from scratch and `IN Sup' refers to ImageNet supervised pre-training. All unsupervised models are pre-trained using COCO \textit{train2017} for 800 epochs.}
\label{tbl:maskrcnn-full}
\end{table}
\begin{table}[!t]
  \begin{center}
  \resizebox{\linewidth}{!}{
    \begin{tabular}{c|ccc|ccc}
    Method & AP$^{bb}$ & AP$^{bb}_{50}$ & AP$^{bb}_{75}$ & AP$^{mk}$ & AP$^{mk}_{50}$ & AP$^{mk}_{75}$  \\
    \hline
    Rand Init &17.7 & 31.0 & 17.7 & 16.5 & 28.8 & 16.6 \\
    IN Sup& 25.5 & 42.4 & 27.0 & 23.3 & 39.4 & 24.3 \\
    \hline
    SimCLR & 24.1 & 40.0 & 25.3 & 22.0 & 37.2 & 22.8 \\
    BYOL & 24.2 & 41.6 & 24.9 & 22.3 & 38.5 & 22.9  \\
    SimSiam & 24.5 & 40.5 & 26.0 & 22.4 & 37.8 & 23.4 \\
    MoCo v2  & 24.2 & 40.1 & 25.5 & 22.2 & 37.6 & 22.9 \\
    DenseCL & 26.1 & 42.7 & 27.2 & 23.8 & 40.1 & 24.9 \\
    \hline
    \ourmethod~(Ours) & \textbf{26.4} & \textbf{43.3} & \textbf{27.9} & \textbf{24.0} & \textbf{40.4} & \textbf{25.1}
†    \end{tabular}}
  \end{center}
  \vspace{-0.4cm}
  \caption{Semi-supervised object detection and instance segmentation on COCO using Mask R-CNN FPN. All models are fine-tuned for 27k iterations using only 10\% of the COCO \textit{train2017} annotated data. }
\label{tbl:maskrcnn-semi}
\end{table}

\noindent  \textbf{Selective search and data augmentation.} 
Selective search~\cite{felzenszwalb2004efficient} is performed offline before pre-training to produce bounding box proposals for each image. Boxes whose IoU are larger than 0.5 are merged. We remove bounding boxes whose areas are smaller than $12^2$ and whose aspect ratios are outside of the range $(0.33, 3)$. During training, images are augmented as in MoCo v2: a random patch of the image is cropped and resized to 224$\times$224, followed by colour distortion, random Gaussian blur, random grayscale and random horizontal flip operations. If after augmentation, a bounding box is clipped by more than 40\% of its original area, it is removed.

\begin{table*}[t!]
  \begin{center}
    \begin{tabular}{c|ccc|ccc|ccc}
           & \multicolumn{3}{c|}{COCO RetinaNet} 
           & \multicolumn{3}{c|}{PascalVOC Faster R-CNN} 
           & \multicolumn{3}{c}{COCO Keypoint R-CNN}\\
    \hline
    Method & AP$^{bb}$ & AP$^{bb}_{50}$ & AP$^{bb}_{70}$ & AP$^{bb}$ & AP$^{bb}_{50}$ & AP$^{bb}_{75}$&  AP$^{kp}$ & AP$^{kp}_{50}$ & AP$^{bb}_{kp}$\\
    \hline
    Rand Init & 25.5 & 40.9 & 26.7 & 39.8 & 65.7 & 41.4 & 63.0 & 85.2 & 69.0 \\
    IN Sup& 37.9 & 57.9 & 40.4 & 54.4 & 80.4 & 59.7 & 65.5 & 87.1 & \textbf{71.8} \\
    \hline
    SimCLR & 36.4 & 55.8 & 38.4 & 53.1 & 78.9 & 58.4 & 65.2 & 86.8 & 70.7  \\
    BYOL & 36.7 & 56.1 & 39.3 & 52.4 & 79.6 & 57.1 & 65.1 & 86.6 & 70.8 \\
    SimSiam & 36.9 & 56.6 & 39.1 & 53.2 & 79.3 & 58.3 & 65.6 & 86.8 &71.5 \\
    MoCo v2 & 36.6 & 56.3 & 38.8 & 53.6 & 79.9 & 58.9 & 65.3 & 86.8 & 71.3  \\
    DenseCL & 38.3 & 58.1 & 41.0 & 55.3 & 80.5 & 60.8 & 65.5 & 87.2 & 71.5  \\
    \hline
    \ourmethod~(Ours) & \textbf{38.4} & \textbf{58.4} & \textbf{41.2} & \textbf{55.8} & \textbf{80.7} & \textbf{61.9} & \textbf{65.7} & \textbf{87.4} & 71.6  \\
    \end{tabular}
  \end{center}
  % \vspace{-0.7cm}
  \caption{Transfer learning performance of COCO object detection using RetinaNet,  Pascal VOC object detection using Faster R-CNN FPN, and COCO Keypoint detection using Keypoint R-CNN FPN. All models are pre-trained on COCO \textit{train2017} for 800 epochs.}
\label{tab:retina-pascal-keypoint}
\end{table*}

\noindent  \textbf{Optimisation.} 
We closely follow the pre-training settings in ~\cite{densecl}. We train the model for 800 epochs with an initial learning rate of 0.3, which is cosine-annealed~\cite{sgdr} to zero. We use an SGD optimiser with weight decay and momentum of 0.0001 and 0.9 respectively. The margin parameter $\alpha$ in Equation~\ref{eq:imdiscrim} is set to $0.4$. We use 8 Nvidia 2080Ti GPUs for pre-training with a mini-batch size of 256. 

\noindent \textbf{Implementation.}
Our code is based on OpenSelfSup\footnote{\href{https://github.com/open-mmlab/OpenSelfSup}{https://github.com/open-mmlab/OpenSelfSup}} and is available at ~\url{https://github.com/ChenhongyiYang/CCOP}. We performed pre-training for competing methods locally\footnote{The exception to this is for Self-EMD~\cite{selfemd} as we could not find any source code. The numbers for Self-EMD that appear in the sequel are as presented in the original paper.} across 8 Nvidia 2080Ti GPUs, using the training schemes given in their respective papers, and the config files in the OpenSelfSup toolkit where available. Downstream models in Section~\ref{sec:mainresults} were trained locally using Detectron2~\cite{wu2019detectron2}.

\subsection{Main Results}
\label{sec:mainresults}

Here, we apply our pre-trained model to a suite of downstream object-level tasks and compare against existing pre-training approaches.

\noindent  \textbf{COCO object detection and segmentation.} 
In Table~\ref{tbl:maskrcnn-full}, we compare the performance of our approach to existing pre-training schemes by fine-tuning a Mask R-CNN FPN model. We followed the training settings in ~\cite{densecl}: the model is trained on the COCO \textit{train2017} subset and tested on the~\textit{mini-val} subset. Synchronised batch normalisation (SyncBN) is used in the backbone, FPN, and detection heads. ~\ourmethod~achieves a 1.4 and 1.3 absolute improvement on AP$^{bb}$ and AP$^{mk}$ compared to the MoCo v2 baseline, and surpasses other recent pre-training approaches including those designed for object-level and pixel-level tasks such as DenseCL~\cite{densecl} and Self-EMD~\cite{selfemd}.%s The high AP$^{bb}_{75}$ and AP$^mk_{75}$ for our approach  indicate that it is able to capture detailed fe it can capture more detailed features.

\noindent  \textbf{Semi-supervised detection and segmentation.} 
In Table~\ref{tbl:maskrcnn-semi}, we report the semi-supervised object detection and instance-segmentation performance on COCO where only 10\% of images are used for fine-tuning\footnote{We note that this is not technically semi-supervised learning as it does not utilise unlabelled data points, but we use the term to be consistent with several of the works we compare to.}. This experiment allows us to gauge how well~\ourmethod~works in a regime where we have limited downstream annotation. Our approach achieves a 2.4 AP$^{bb}$ and 1.8 AP$^{mk}$  improvement over the baseline MoCo v2 model and is state-of-the-art.

\noindent  \textbf{COCO single-stage object detection.} 
Here, we show that the model pre-trained using \ourmethod~can be used in a single stage object detector, demonstrating its flexibility. We transfer the pre-trained model to a RetinaNet~\cite{lin2017focal} detector using 1$\times$ training schedule. SyncBN is used in the backbone and FPN, and group normalisation is used in the detection heads. Results are given in Table~\ref{tab:retina-pascal-keypoint}. Our model attains 1.8 absolute AP$^{bb}$ improvement compared to the baseline of MoCo v2, which is larger than the improvement for the two-stage Mask R-CNN. This could be because our method is able to pre-train not only the backbone network, but also the FPN neck that plays an more important role in single stage detectors~\cite{li2019scale}.

\noindent  \textbf{Pascal VOC object detection.} 
To verify whether our model can generalise to datasets other than COCO, we transfer the pre-trained model to a Faster R-CNN FPN detector and fine-tune it using the Pascal VOC~\cite{everingham2015pascal} dataset. Following~\cite{moco}, we train the model for 24k iterations using the \textit{trainval2007+2012} sets and evaluate it on the \textit{test2007} set. The results are provided in Table~\ref{tab:retina-pascal-keypoint}; our methods achieves a 2.2 improvement in AP over the baseline and surpasses other state-of-the-art approaches, indicating that our model is able to generalise to new data.

%Similar to COCO, we notice that our method can improve AP$_{75}$ by a large margin, suggesting it can help with learn more spatial-sensitive features.

\noindent  \textbf{COCO human keypoint detection.} 
Our pre-trained model also works well when used downstream for human keypoint detection. In Table~\ref{tab:retina-pascal-keypoint}, we provide the results obtained when adapting our pre-trained model to a Keypoint R-CNN FPN~\cite{he2017mask} model. Our approach achieves 65.7 AP$^{kp}$, broadly improving on alternative pre-training methods.

\subsection{Ablation Studies}
In this section, we examine how the different aspects of~\ourmethod~contribute to the overall performance. We use COCO object detection and instance segmentation to evaluate different models, which are fine-tuned using a Mask R-CNN FPN model with 1$\times$ training schedule.

\begin{table}[t]
  \begin{center}
  \resizebox{\linewidth}{!}{
    \begin{tabular}{ccc|ccc|ccc}
    Obj & Intra & \pcurrshort & AP$^{bb}$ & AP$^{bb}_{50}$ & AP$^{bb}_{75}$ & AP$^{mk}$ & AP$^{mk}_{50}$ & AP$^{mk}_{75}$ \\
    \hline
    & & & 38.5 & 58.0 & 41.9 & 34.9 & 55.2 & 37.3 \\
  \checkmark     &  & & 39.3 & 59.3 & 42.9 & 35.6 & 56.5 & 38.2 \\
  \checkmark    & \checkmark  & & 39.6 & 59.5 & 43.2 & 35.8 & 56.2 & 38.5 \\
  \checkmark    &   & \checkmark & 39.7 & \textbf{59.9} & 43.4 & 36.0 & \textbf{56.9} & 38.5 \\
  \checkmark    & \checkmark  & \checkmark & \textbf{39.9} & 59.8 & \textbf{43.9} & \textbf{36.2} & 56.8 & \textbf{38.8} \\
    \end{tabular}}
  \end{center}
  \vspace{-0.4cm}
  \caption{Ablation Studies of our contributions. `Obj' stands for object-level contrastive learning; `Intra' stands for intra-image discrimination loss; \pcurrshort~is our~\pcurrfull~approach.}
\label{tbl:ablation-study}
\end{table}
\begin{table}[t]
  \begin{center}
  \resizebox{\linewidth}{!}{
    \begin{tabular}{c|ccc|ccc}
    Method &AP$^{bb}$ & AP$^{bb}_{50}$ & AP$^{bb}_{75}$ & AP$^{mk}$ & AP$^{mk}_{50}$ & AP$^{mk}_{75}$   \\
    \hline
    MoCo v2 & 38.5 & 58.0 & 41.9 & 34.9 & 55.2 & 37.3 \\
    Random & 39.1 & 58.9 & 42.8 & 35.1 & 55.8 & 37.8 \\
    GT & 39.4 & 59.3 & 43.2 & 35.5 & 56.4 & 38.2 \\
    SS & \textbf{39.9} & \textbf{59.8} & \textbf{43.9} & \textbf{36.2} & \textbf{56.8} & \textbf{38.8} \\
    \end{tabular}}
  \end{center}
  \vspace{-0.4cm}
  \caption{Comparison of different object generation approaches. `Random' stands for random generated boxes; `GT' stands for human annotated boxes; `SS' stands for selective search.}
\label{tbl:boxgen}
\end{table}

\begin{table}[t]
  \begin{center}
  \resizebox{\linewidth}{!}{
    \begin{tabular}{c|ccc|ccc}
    $\beta$ &AP$^{bb}$ & AP$^{bb}_{50}$ & AP$^{bb}_{75}$ & AP$^{mk}$ & AP$^{mk}_{50}$ & AP$^{mk}_{75}$   \\
    \hline
    1.0 & 39.6 & 59.5 & 43.2 & 35.8 & 56.2 & 38.5 \\
    0.6 & 39.5 & 56.5 & 42.9 & 35.7 & 56.5 & 38.0 \\
    0.3 & 39.2 & 58.9 & 42.7 & 35.5 & 56.1 & 38.2 \\
    \pcurrshort & \textbf{39.9} & \textbf{59.8} & \textbf{43.9} & \textbf{36.2} & \textbf{56.8} & \textbf{38.8} \\
    \end{tabular}}
  \end{center}
  \vspace{-0.4cm}
  \caption{Comparison of box augment strategies with different IoU lower-bound $\beta$. Note that 1.0 stands for no augmentation. }
\label{tbl:curr}
\end{table}

\noindent  \textbf{Role of each module.} 
In Table~\ref{tbl:ablation-study} we examine how object-level contrastive learning, the intra-image  discrimination loss, and \pcurrfull~contribute to the overall performance. Our baseline MoCo v2 model achieves 38.5 AP$^{bb}$ and 34.9 AP$^{mk}$. When we include the object-level contrastive learning loss, the performance increases significantly to 39.3 AP$^{bb}$ and 35.6 AP$^{mk}$. This shows that incorporating object-level information in the pre-training stage can help with downstream object-level tasks. When we add the intra-image discrimination loss, the performance is further enhanced to 39.6 AP$^{bb}$ and 35.8 AP$^{mk}$. Separately adding \pcurrshort~without the intra-image loss also improves performance to 39.7 AP$^{bb}$ and 36.0 AP$^{mk}$. Finally, when all three modules are adopted, we achieve 39.9 AP$^{bb}$ and 36.2 AP$^{mk}$. 

% which validate our assumption that gradually making the positive pairs harder to be matched is good for learning robust regional features. Finally, when all the three things are applied, the final performance achieves 39.9 AP$^{bb}$ and 36.2 AP$^{mk}$.   

\noindent  \textbf{Object Generation.} 
We compare proposing regions using selective search to two alternative strategies: randomly generating proposals, and using ground-truth bounding boxes. For random generation, we produce 12 bounding boxes per view (this is on average how many boxes selective search produces) and we remove boxes whose areas are smaller than $12^2$ and whose aspect ratios are out of range $(0.33, 3)$. For ground-truth, we use the COCO annotations. The performance for these strategies is given in Table~\ref{tbl:boxgen}. Using randomly generated boxes gives us a marginal improvement over the baseline but is worse than using ground-truth boxes or those obtained through selective search. Our intuition is that random boxes allow the model to learn spatially-sensitive features, but do not give any notion of what an object is. We found it intriguing that selective search boxes outperform ground-truth. A possible explanation is that selective search introduces more diversity into training.

\begin{table}[t]
  \begin{center}
  \resizebox{\linewidth}{!}{
    \begin{tabular}{c|ccc|ccc}
    $\alpha$ &AP$^{bb}$ & AP$^{bb}_{50}$ & AP$^{bb}_{75}$ & AP$^{mk}$ & AP$^{mk}_{50}$ & AP$^{mk}_{75}$   \\
    \hline
    0.0 & 38.8 & 58.1 & 41.9 & 35.2 & 55.0 & 37.4 \\
    0.2 & 39.5 & 59.5 & 42.9 & 35.6 & 56.5 & 38.0 \\
    0.4 & \textbf{39.9} & \textbf{59.8} & \textbf{43.9} & \textbf{36.2} & \textbf{56.8} & \textbf{38.8} \\
    0.6 & \textbf{39.9} & \textbf{59.8} & 43.4 & 36.0 & \textbf{56.8} & 38.5 \\
    \end{tabular}}
  \end{center}
  \vspace{-0.4cm}
  \caption{Ablation study of the margin parameter $\alpha$ in the intra-image discrimination loss.}
\label{tbl:margin}
\end{table}

\begin{table}[t]
  \begin{center}
  \resizebox{\linewidth}{!}{
    \begin{tabular}{c|c|ccc|ccc}
    Backbone & Method &AP$^{bb}$ & AP$^{bb}_{50}$ & AP$^{bb}_{75}$ & AP$^{mk}$ & AP$^{mk}_{50}$ & AP$^{mk}_{75}$ \\
    \hline
    \multirow{2}{*}{ResNet-18} & MoCo v2 & 35.0 & 53.9 & 38.2 & 31.8 & 51.2 & 34.0\\
     & \ourmethod & 35.6 & 54.7 & 38.6 & 32.3 & 52.0 & 34.4\\
    \hline
    \multirow{2}{*}{ResNet-50} & MoCo v2 & 38.5 & 58.0 & 41.9 & 34.9 & 55.2 & 37.3\\
     & \ourmethod & 39.9 & 59.8 & 43.9 & 36.2 & 56.8 & 38.8\\
    \hline
    \multirow{2}{*}{ResNet-101} & MoCo v2 & 41.0 & 60.7 & 44.5 & 36.7 & 57.6 & 39.5\\
     & \ourmethod & 42.1 & 61.9 & 45.9 & 37.6 & 58.7 & 40.3\\ 
    \end{tabular}}
  \end{center}
  \vspace{-0.4cm}
  \caption{Performance comparison between our \ourmethod~and the baseline MoCo v2 on different backbone sizes.}
\label{tbl:modelsize}
\end{table}

\noindent  \textbf{Curriculum Learning.}
We conduct experiments to study the extent to which curriculum learning benefits our approach. We compare \pcurrshort~to box generations with fixed magnitude noise and with a fixed IoU lower-bound $\beta$. The results are presented in Table~\ref{tbl:curr}. Note that when $\beta=1$ there is no augmentation applied. We observe that even if no noise is added, the pre-trained model is quite good achieving a 39.6 AP$^{bb}$ and 35.8 AP$^{mk}$. Lowering the IoU lower-bound to 0.6 barely changes the performance. However when we lower it to 0.3 this harms performance. We observe that \pcurrfull~can avoid such problems by gradually enlarging the extent of augmentation. This stabilises training for the early stages of learning and raise the difficulty in later stages, leading to a superior performance of 39.9 AP$^{bb}$ and 36.2 AP$^{mk}$.

\noindent  \textbf{Intra-image Margin}
In Table~\ref{tbl:margin} we explore the influence of the margin $\alpha$ parameter in the  intra-image discrimination loss. We can see that the performance is robust for large margins, but low for small values. A possible reason is that a small margin makes it difficult for the model to learn good embeddings as it introduce too much conflict in the image.

\begin{figure}[!t]
    \includegraphics[width=\linewidth]{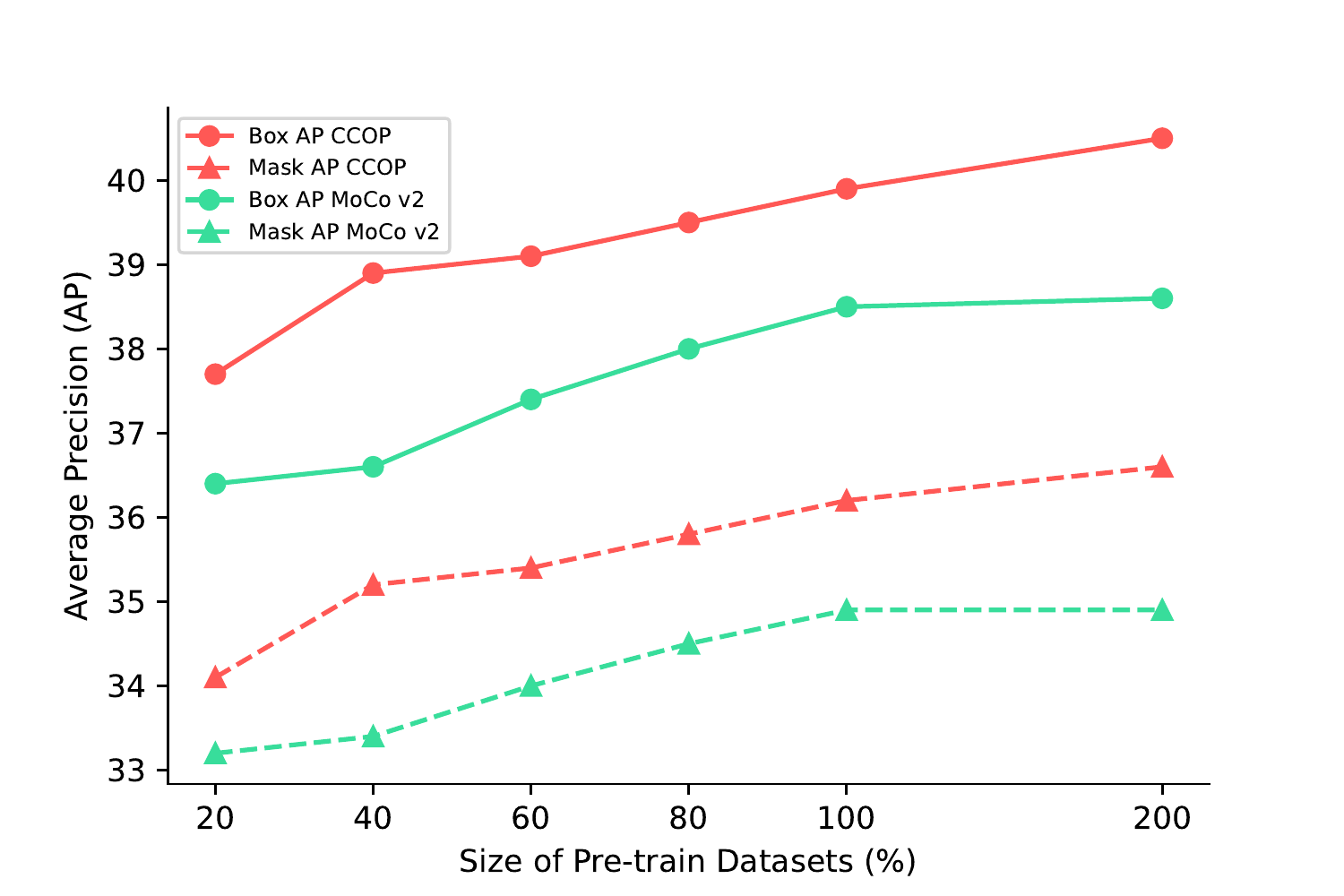}
    % \vspace{-0.5cm}
    \caption{Performance comparison between the baseline MoCo v2 and our \ourmethod~on Mask R-CNN FPN with 1$\times$ training schedule under different sizes of pre-training dataset. Note that 200\% means that we add the extra COCO \textit{unlabeled} set for pre-training. As we increase the dataset size, CCOP performance continues to increase.}
    \label{fig:size}
\end{figure}

\noindent  \textbf{Model sizes.}
In Table~\ref{tbl:modelsize}, we compare the performance of~\ourmethod~and the baseline MoCo v2 using ResNets of different sizes. There are performance boosts when using our approach in all cases. However, these are more significant as the models get larger.

\noindent  \textbf{Pre-train dataset size.}
In Figure~\ref{fig:size}, we show the performance of our approach and the baseline model for different sizes of the pre-training dataset. We take subsets with different percentages of the COCO training set and use them to pre-train the model for 800 epochs. We also test the performance when we add the COCO \textit{unlabeled} set into pre-training, which contains an extra 123k images. We observe that the performance of the baseline starts to saturate earlier than our method. This suggests that our method has the potential for large-scale pre-training.

\noindent  \textbf{Pre-train dataset size.}
In Figure~\ref{fig:size}, we show the performance of our approach and the baseline model for different sizes of the pre-training dataset. We take subsets with different percentages of the COCO training set and use them to pre-train the model for 800 epochs. We also test the performance when we add the COCO \textit{unlabeled} set into pre-training, which contains an extra 123k images. We observe that the performance of the baseline starts to saturate earlier than our method. This suggests that our method has the potential for large-scale pre-training.

\subsection{Qualitative Studies}
We qualitatively study how~\ourmethod~learns useful features from bounding boxes generated in an unsupervised manner on multi-object input images and how this helps with downstream object-level tasks.

\begin{figure}[!t]
    \begin{subfigure}[b]{0.48\linewidth}
    \includegraphics[width=\textwidth]{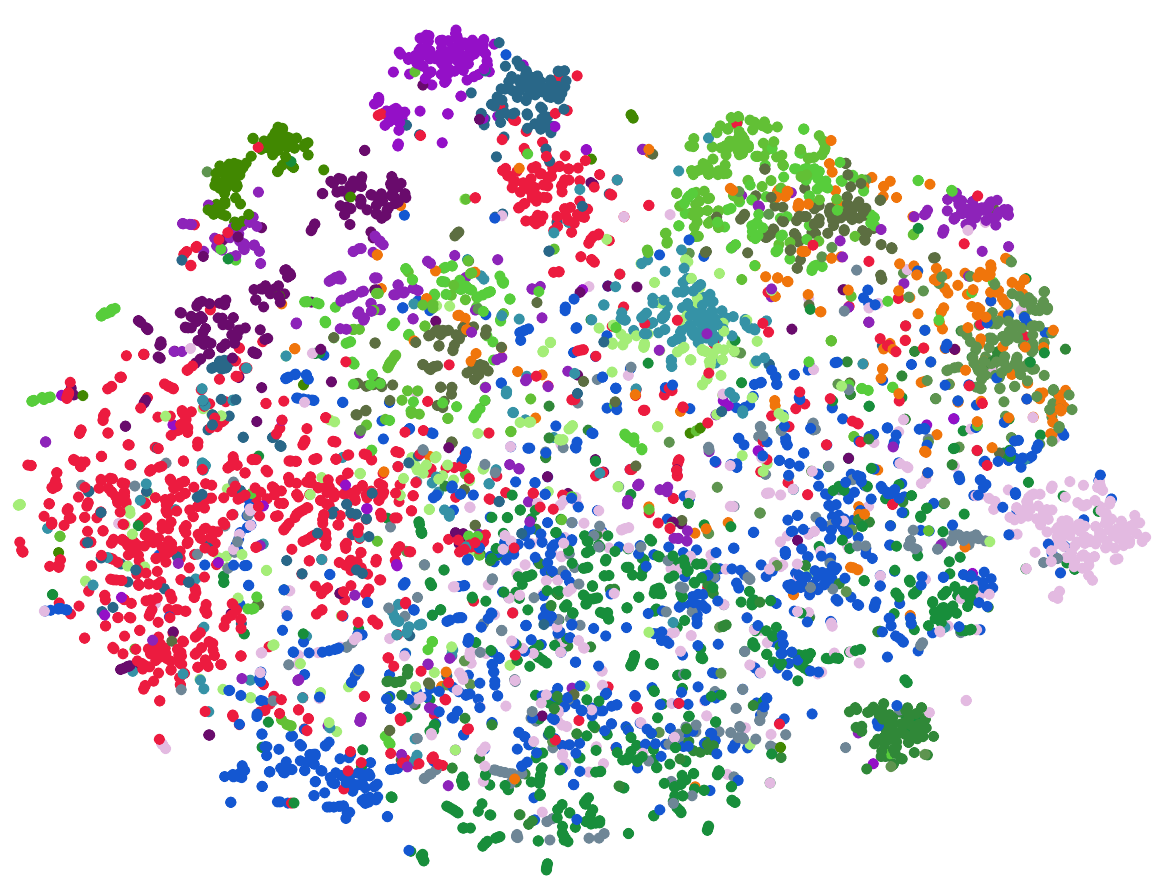}
    \caption{Backbone}
    \end{subfigure} 
    \begin{subfigure}[b]{0.48\linewidth}
    \includegraphics[width=\textwidth]{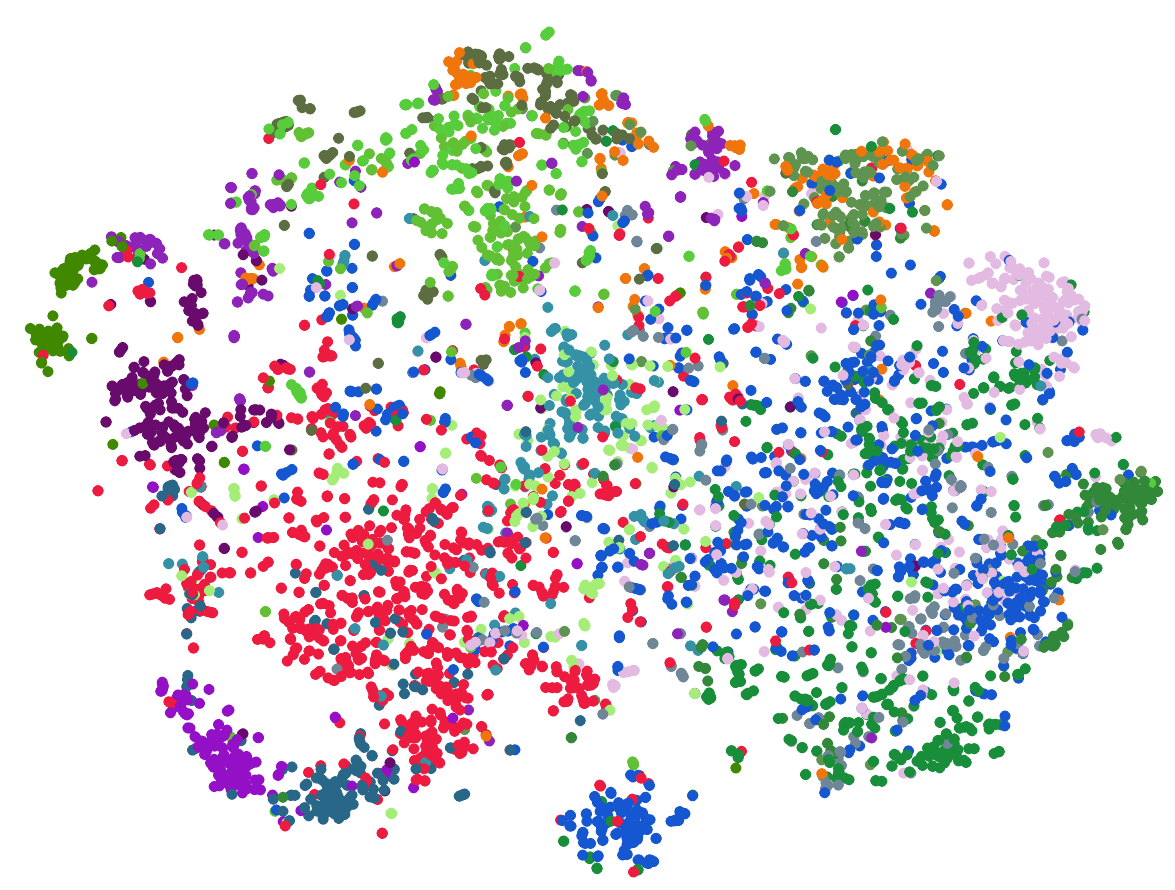}
    \caption{Prediction}
    \end{subfigure}
    \caption{Visualisation of regional backbone features and predicted box embeddings on COCO \textit{mini-val} set using T-SNE for dimension reduction. The objects are acquired by using oracle annotations. We only draw the common 20 classes, shown in different colors, that appears in both COCO and Pascal VOC.}
    \label{fig:embedding}
\end{figure}

In Figure~\ref{fig:embedding}, we visualise the learnt object embeddings using T-SNE~\cite{tsne}. In detail, we forward the COCO \textit{mini-val} images through the pre-trained model and compute the object embeddings of the ground-truth boxes, then pick 20 common classes (denoted with different colours) that appear in both COCO and Pascal VOC for visualisation. We can see that the box embeddings in the same class form a series of clear clusters in the embedding space even though the model is trained without supervision, demonstrating the model's ability to capture high-level semantic information. As only the pre-trained backbone is used for fine-tuning, we also visualise the regional features that are directed pooled and averaged from the backbone network. The clustering structure can also be clearly observed in these features even before they are transformed using an MLP head. Further, we query the $k$ nearest neighbours of each object in the representation space and check how many of them are in the same class, which we denote as kNN recall. These values are given in Table~\ref{tbl:recall}. The performance using backbone features is only slightly worse than the performance using the actual embeddings, indicating that the pre-trained model has strong potential for transfer learning.
\begin{table}[!t]
  \begin{center}
    \begin{tabular}{c|cccccc}
    Topk & 1 & 5 & 10 & 20 & 50 & 100  \\
    \hline
    Prediction & 50.1 & 44.4 & 40.8 & 37.5 & 33.6 & 30.5 \\
    Backbone & 48.1 & 43.2 & 39.4 & 35.9 & 31.7 & 28.3 \\
    \end{tabular}
  \end{center}
  \vspace{-0.4cm}
  \caption{Comparison of kNN recall between the predicted object embeddings and the backbone regional features on COCO \textit{mini-val} set using oracle boxes and class annotations.}
  % \vspace{-0.4cm}
\label{tbl:recall}
\end{table}

\begin{figure}[!t]
    \includegraphics[width=\linewidth]{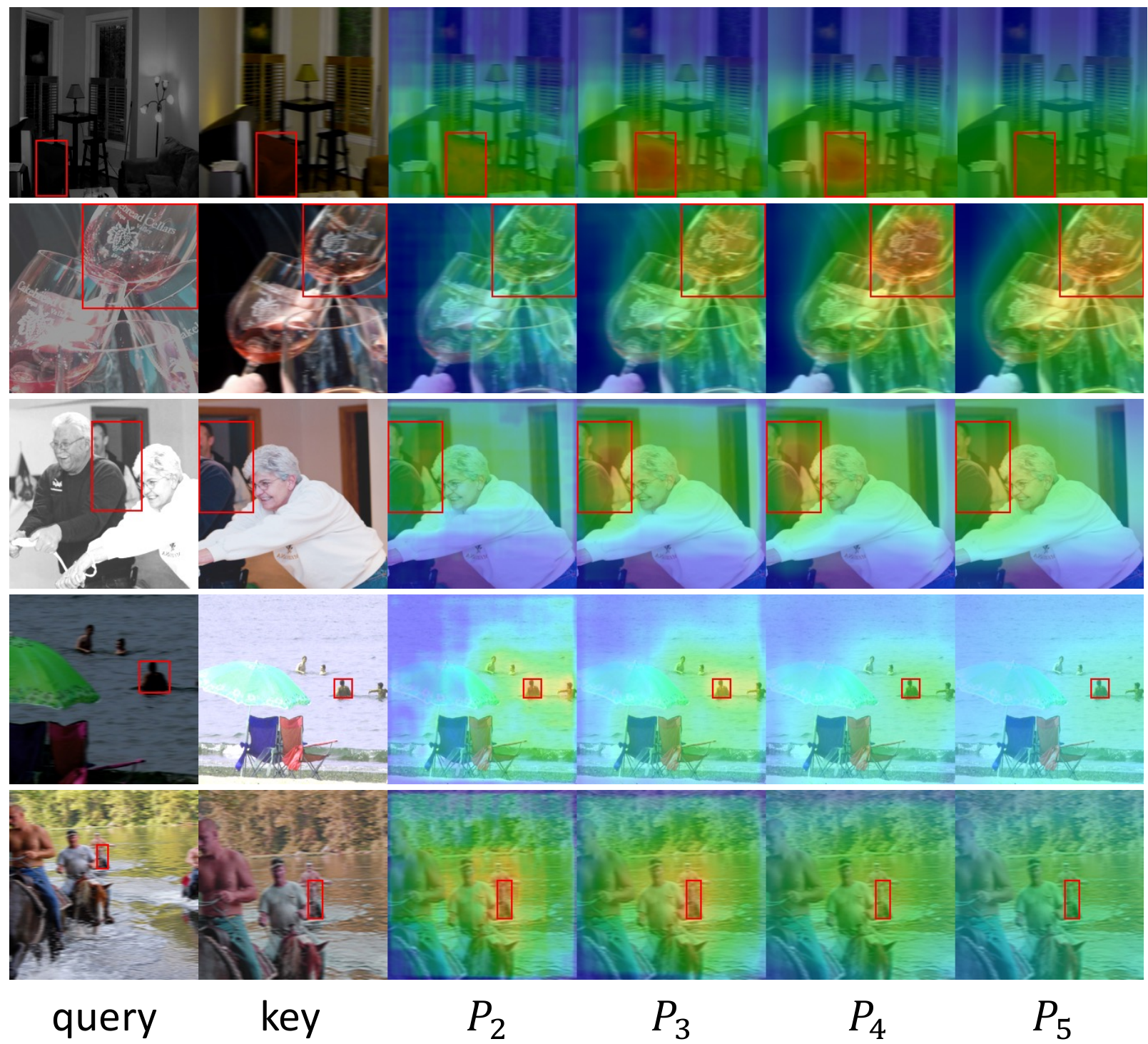}
    \caption{Visualisation of box query results. We pool the box features directly from the query network's backbone features and used it as convolution kernel to compute the cross-correlation results with key image's features in different FPN layers.}
    \label{fig:query}
    % \vspace{-0.2cm}
\end{figure}

Finally, we observe how good our approach is at learning spatially-sensitive and scale-sensitive features that are important to object-level tasks. First, two augmented views of an image are fed into the query and key branch respectively. Then we pool and average an object's regional features from the query network and use this it as a convolution kernel to compute the cross-correlation with the backbone features computed in the key branch. The results are in Figure~\ref{fig:query}. We can see that the same object in the key image is highlighted; this suggests that the pre-trained model is able to extract features that are suitable for object-level tasks.

\section{Conclusion}
In this work, we propose an object-level contrastive pre-training framework:~\ourmethodFull~(CCOP)~to improve transfer learning performance for downstream object-level tasks. Our approach is designed for pre-training on multi-object scene images that take both inter-image and intra-image instance discrimination into consideration. To further improve the performance, we designed \pcurrfull~that adaptively selects the positive pairs of an appropriate difficulty for the current stage of training. Extensive experimental results verify that a model pre-trained using our method is able to transfer well to different tasks, different downstream model architectures, and different datasets.

\paragraph{Acknowledgements.} 

The authors would like to thank
Adam Jelly,
Antreas Antoniou,
Jiawei He,
Joe Mellor,
Justin Engelmann,
and Zehui Chen 
for their helpful feedback and suggestions. Chenhongyi Yang was supported by a PhD studentship provided by the School of Engineering, University of Edinburgh.

{\small
\bibliographystyle{ieee_fullname}
\bibliography{egbib}
}

\end{document}